\documentclass[conference]{IEEEtran}
\IEEEoverridecommandlockouts
\usepackage{cite}
\usepackage{amsmath,amssymb,amsfonts}
\usepackage{algorithmic}
\usepackage{graphicx}
\usepackage{textcomp}
\usepackage{xcolor}

\usepackage{subfigure}
\usepackage{bbm}
\usepackage{multirow}
\usepackage{booktabs}

\def\BibTeX{{\rm B\kern-.05em{\sc i\kern-.025em b}\kern-.08em
    T\kern-.1667em\lower.7ex\hbox{E}\kern-.125emX}}

\begin{document}

\title{General GAN-generated Image Detection by Data Augmentation in Fingerprint Domain}

\author{\IEEEauthorblockN{Huaming Wang$^{\ast}$\thanks{{$^{\ast}$}Equal contribution.}}
\IEEEauthorblockA{\textit{College of Cyber Security} \\
\textit{Jinan University}\\
Guangzhou, China \\
w778741197@163.com}
\and
\IEEEauthorblockN{Jianwei Fei$^{\ast}$}
\IEEEauthorblockA{\textit{School of Computer Science} \\
\textit{Nanjing University of Information Science and Technology}\\
Nanjing, China \\
fjw826244895@163.com}
\and
\IEEEauthorblockN{Yunshu Dai}
\IEEEauthorblockA{\textit{School of Computer Science} \\
\textit{Nanjing University of Information Science and Technology}\\
Nanjing, China \\
daiyunshu0102@163.com}
\and
\IEEEauthorblockN{Lingyun Leng}
\IEEEauthorblockA{\textit{College of Cyber Security} \\
\textit{Jinan University}\\
Guangzhou, China \\
portgasly@163.com}
\and
\IEEEauthorblockN{Zhihua Xia\dag\thanks{{\dag}Corresponding author.}}
\IEEEauthorblockA{\textit{College of Cyber Security} \\
\textit{Jinan University}\\
Guangzhou, China \\
 xia\_zhihua@163.com}
}

\maketitle

\begin{abstract}
In this work, we investigate improving the generalizability of GAN-generated image detectors by performing data augmentation in the fingerprint domain.
Specifically, we first separate the fingerprints and contents of the GAN-generated images using an autoencoder based GAN fingerprint extractor, followed by random perturbations of the fingerprints.
Then the original fingerprints are substituted with the perturbed fingerprints and added to the original contents, to produce images that are visually invariant but with distinct fingerprints.
The perturbed images can successfully imitate images generated by different GANs to improve the generalization of the detectors, which is demonstrated by the spectra visualization.
To our knowledge, we are the first to conduct data augmentation in the fingerprint domain.
Our work explores a novel prospect that is distinct from previous works on spatial and frequency domains augmentation.
Extensive cross-GAN experiments demonstrate the effectiveness of our method compared to the state-of-the-art methods in detecting fake images generated by unknown GANs.
\end{abstract}


% -----------------------------换节分隔符-----------------------------

\section{Introduction}
\begin{figure}
    \centering
    \subfigure[Before fingerprint domain augmentation]{
    \label{Fig1.sub1}
    \includegraphics[width=0.98\linewidth]{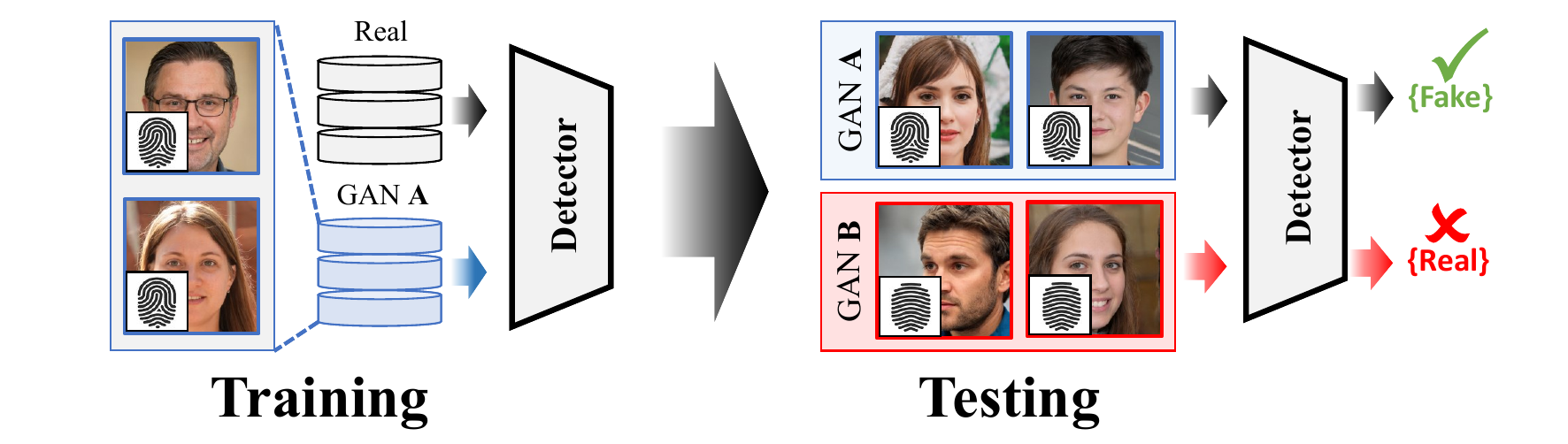}}
        
    \subfigure[After fingerprint domain augmentation]{
    \label{Fig1.sub2}
    \includegraphics[width=0.98\linewidth]{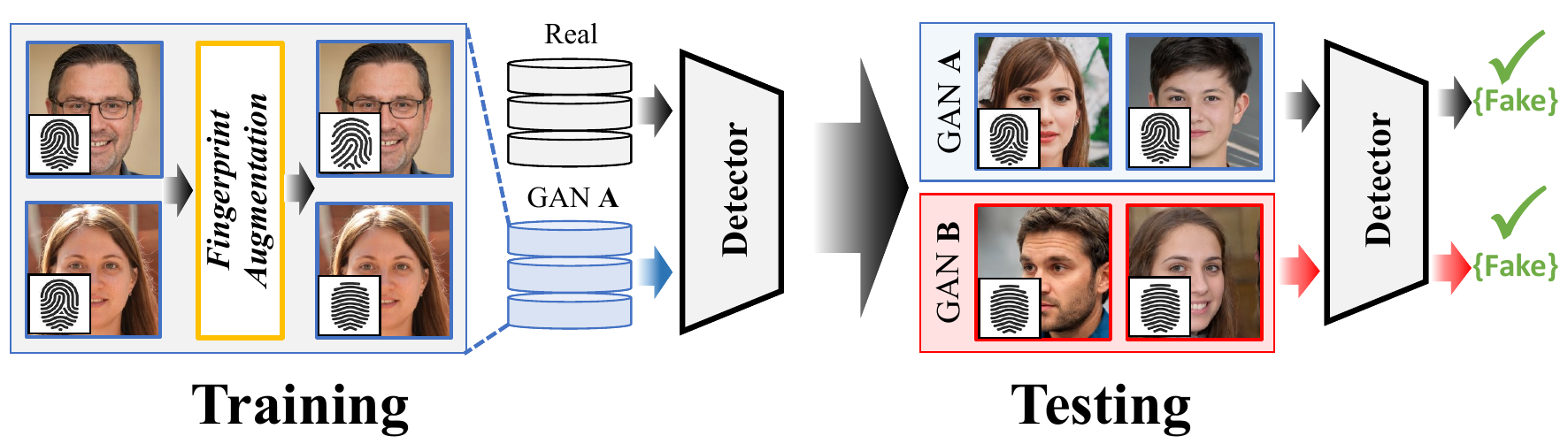}}
    
    \caption{Illustration of how the proposed GAN fingerprint augmentation influences the detection performance of unseen GANs.
    Zoom in for a better view.
    The fingerprint pattern on the fake image signifies such a GAN fingerprint is possessed by that image.
    The upper figure (a) shows that the detector trained on GAN \textit{A} cannot be generalized to GAN \textit{B} well, while (b) shows that our method enables the detector to train on fake images with wider distribution of GAN fingerprints, to realize better generalizability.}
\label{fig:fingerprints}
\end{figure}

Recent advancements in generative adversarial networks (GANs)~\cite{goodfellow2020generative, karras2018progressive, karras2020analyzing} enable them to generate highly photorealistic fake images that are indistinguishable from naked eyes, resulting in a potential threat of malicious misuse to individuals and society.
As a result, many methods have been proposed to identify GAN-generated images and achieved promising performances.
However, generalizing to images generated by unknown GANs remains a great challenge for the detectors.
Most existing works~\cite{zhang2019detecting, wang2020cnn, durall2020watch, frank2020leveraging, jeong2022bihpf} have achieved flawless detection accuracy on images from the same distribution as the training data, i.e., generated by the same GANs, but suffer from a significant drop in images generated by unseen GANs, as shown in Fig.~\ref{Fig1.sub1}.
Although some recent forensic works~\cite{fei2022learning, dai2022attentional} based on local anomaly have made efforts to improve the generalization of face forgery detection, they are not so applicable to GAN-generated image detection.

Existing works have demonstrated that GAN-generated image detectors are inclined to distinguish real images and GAN-generated images by a subtle but general trace~\cite{neves2020ganprintr}.
This trace, referred to as \textit{GAN fingerprint}, is an invisible pattern that the GAN leaves in the images it generates.
GAN fingerprint is the most discriminative feature~\cite{marra2019gans, yu2019attributing} caused by up-sampling and varies from different GAN architectures.
The detectors have poor performance on images generated by unseen GANs since the fingerprints are distinct.
In other words, the performance degradation of cross-GAN evaluations is due to the overfitting of GAN-specific fingerprints.

In this paper, we propose to improve the cross-GAN performance of the detectors by performing data augmentation in the fingerprint domain.
The key idea behind our work is based on the fact that the detectors distinguish real and fake images by the GAN fingerprints, while augmenting the fingerprints in the training data can enrich the fake types seen by the detectors.
Unlike prior works that conduct data augmentation in the spatial domain, the augmentation in our method is performed directly in the fingerprint domain.
Specifically, we use an autoencoder based fingerprint extractor to process the fake images and extract the fingerprints by subtracting the reconstructed images from the original images.
Then, the extracted fingerprints are randomly perturbed to simulate the fingerprints of other unknown GAN architectures.
Finally, the original fingerprints of the fake images are replaced by the perturbed fingerprints to obtain augmented fake images.
The detector trained with the fingerprint domain augmentation can learn a wider distribution of fake images, even with images from one single GAN, as shown in Fig.~\ref{Fig1.sub2}.

Our contributions can be summarized as follows:
\begin{itemize}
\item We propose a novel fingerprint domain augmentation method to improve the generalizability of GAN detectors.
By perturbing the fingerprints of fake images using the proposed perturbation strategies, the detectors can be generalized to unseen GAN architectures well.
\item Unlike previous works that perform data augmentation in the spatial domain, our method effectively improves the fundamental deficiency of the detectors that rely on GAN fingerprints in terms of generalizability.
\item Extensive experiments demonstrate the superiority of the proposed fingerprint domain augmentation, and the results show that our augmentation strategies can greatly improve the generalizability of the GAN detectors compared to existing works.
\end{itemize}

% -----------------------------换节分隔符-----------------------------

\section{Related Work}
In this section, we briefly introduce recent works on the detection of GAN-generated images and review some works on GAN fingerprints.

\textbf{GAN Forensics.}
In recent years, GAN forensics have arisen many concerns.
Early studies attempted to distinguish fake images from real images using traces in different color spaces.
McCloskey \textit{et al.}~\cite{mccloskey2019detecting} found that GAN-generated images have statistical anomalies in different color spaces in terms of pixel intensities.
Similarly, Li \textit{et al.}~\cite{li2020identification} suggested a co-occurrences based feature set of different color components to capture color statistics for the detection of GAN-generated images.
Real images and GAN-generated images also show significant differences in the frequency domain.
% Zhang \textit{et al.}~\cite{zhang2019detecting} pointed out the existence of significant artifacts in the spectrum of GAN-generated images and identified the artifacts by simulating such artifacts.
Durall \textit{et al.}~\cite{durall2020watch} pointed out that GANs are not able to reproduce the spectral distribution of real images, and proposed to detect GAN-generated images through azimuthal integration of the Fourier spectrum.
Frank \textit{et al.}~\cite{frank2020leveraging} further analyzed the frequency artifacts of images generated by GANs.
Recently, Jeong \textit{et al.}~\cite{jeong2022frepgan} proposed to enhance the robustness of the detector by adding frequency perturbations to the training data.

\textbf{GAN Fingerprints.}
Previous works have proved the existence and uniqueness of GAN fingerprints and indicated that GAN fingerprints are discriminative clues for the detection of GAN-generated images.
Marra \textit{et al.}~\cite{marra2019gans} first found that GANs leave unique fingerprints in the images they generate.
They used the average noise residuals of the GAN-generated images as fingerprints and proved that they can be used for the detection and attribution of fake images.
Yu \textit{et al.}~\cite{yu2019attributing} further categorized GAN fingerprints into image fingerprints and model fingerprints, and carried out a more comprehensive analysis.
Neves \textit{et al.}~\cite{neves2020ganprintr} proposed a method named GANprintR to remove GAN fingerprints while maintaining the appearance of fake images.
Their experiments showed that the recall rate of the detector decreased dramatically after removing fingerprints from fake images, indicating that the detectors recognized fake images by fingerprints.

In this work, based on the observations of prior works on GAN fingerprints, we directly conduct data augmentation in the fingerprint domain, which leads to a better generalization of the detectors.
Our approach is inherently different from any existing work that aims to improve generalizability.
We explore a new path other than the spatial and frequency domains and advance the GAN fingerprint in GAN forensics and traceability further.

% -----------------------------换节分隔符-----------------------------

\section{Proposed Method}
\begin{figure*}
\centering
\includegraphics[width=0.92\linewidth]{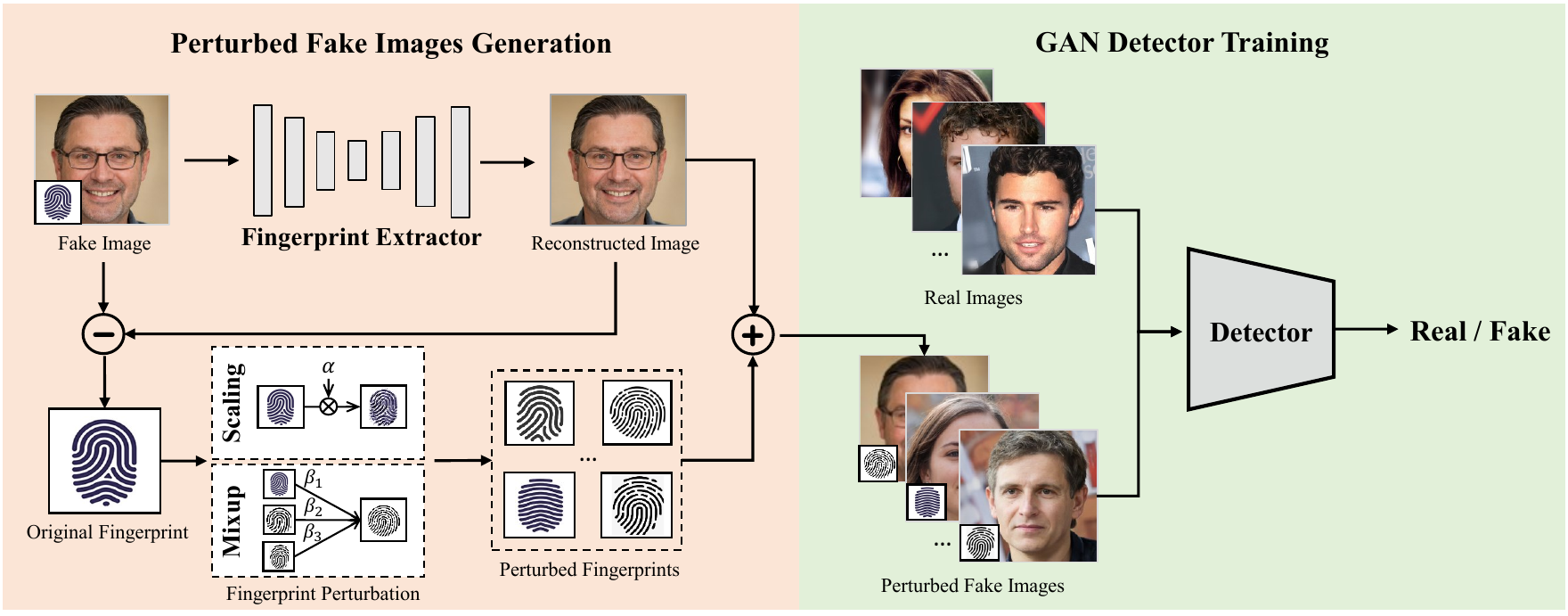}
\caption{Overview of the proposed method.}
\label{fig:overview}
\end{figure*}

As illustrated in Fig.~\ref{fig:overview}, our framework consists of two stages: the perturbed fake images generation and the GAN detector training. 
The former stage enriches the fake types of training data by perturbing the original fingerprints of fake images.
The latter stage trains the GAN detector on real images and perturbed fake images for better generalization.

\subsection{Perturbed Fake Images Generation}

\subsubsection{Fingerprint Extraction}
Autoencoder is able to learn the key structure of real images due to its powerful reconstruction ability.
Thus, an autoencoder trained on real images possesses the prior knowledge of naturalness, and cannot perfectly recover the fake traces in GAN-generated images.
To this end, we propose to extract the GAN fingerprints using an autoencoder mainly trained on real images.
Given that GAN fingerprints are high-frequency discriminative information possessed by only GAN-generated images, we represent the GAN fingerprints as the residuals between the fake images and the reconstructed images.
Let us denote the autoencoder based fingerprint extractor and the real image by $E$, $x_r$ respectively.
In the training phase, $E$ is optimized by minimizing the mean square error (MSE) based reconstruction loss:
\begin{equation}
\setlength{\abovedisplayskip}{5pt}
\setlength{\belowdisplayskip}{5pt}
\mathcal{L}_{rec}=\left\|{x_r}-E(x_r)\right\|_{2}.
\end{equation}

\begin{figure}[t]
\centering
\includegraphics[width=0.85\linewidth]{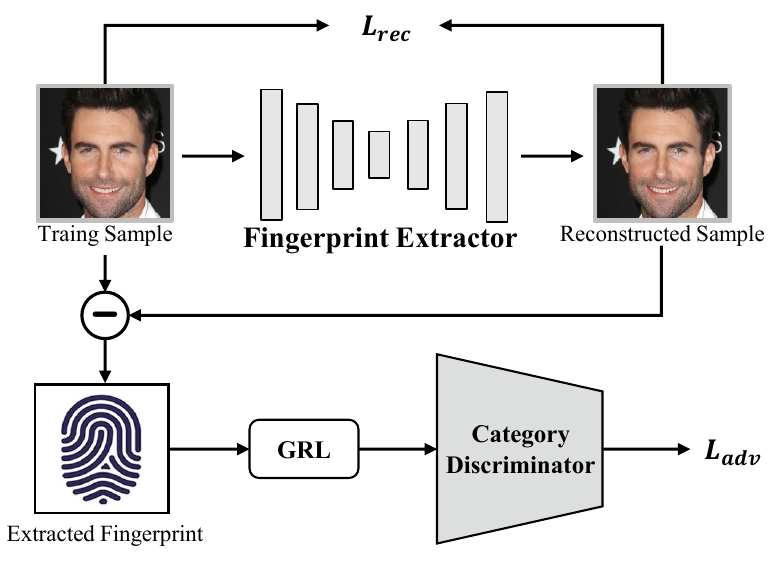}
\caption{Training of the fingerprint extractor.}
\label{fig:extractor}
\end{figure}

However, the indistinguishability of fingerprints from different categories of fake images is not guaranteed.
For example, the fingerprints of fake dogs and fake cats generated by the same GAN may differ, which reduces the universality of the fingerprints.
To further ensure the GAN fingerprints learned by the extractor are content-independent, we leverage an additional category discriminator to make the fingerprints extracted from different categories indistinguishable.
To be specific, besides the MSE on only real images, the fingerprint extractor $E$ is also optimized by maximizing an adversarial loss.
Let us denote the category discriminator by ${D}$, and suppose there are $K$ categories of fake images in the training data, ${D}$ aims to minimizing a $K$-classification loss.
As shown in Fig.~\ref{fig:extractor}, to let the fingerprints indistinguishable for different categories, we utilize a Gradient Reversal Layer (GRL)~\cite{ganin2015unsupervised} between the extracted fingerprints and ${D}$. GRL changes the sign of the gradient from ${D}$ when optimizing ${E}$, to let the extracted fingerprints of different categories indistinguishable.
The training process can be formulated as:
\begin{equation}
\begin{split}
\setlength{\abovedisplayskip}{5pt}
\setlength{\belowdisplayskip}{5pt}
&\min _{D} \max _{E} \mathcal{L}_{adv}(D,E)=\\
&-\mathbbm{E}_{\left(x_f, y\right) \sim \left(X_{f}, Y\right)} \sum_{n=1}^{K} \mathbbm{1}_{[n=y]} \log D\left(x_f-E\left(x_f\right)\right),\\
\end{split}
\end{equation}
where $X_{f}$ and $Y$ are the set of fake images and their corresponding category labels. $\mathbbm{1}_{[n=y]}$ is a one-hot vector with $y$-th element being 1. Outputs of $D(\cdot)$ is a $K$-dimensional vector indicating the predicted category by the index of largest element.

Finally, the total loss $\mathcal{L}_{E}$ of the fingerprint extractor is:
\begin{equation}
\setlength{\abovedisplayskip}{5pt}
\setlength{\belowdisplayskip}{5pt}
\mathcal{L}_{E}=\mathcal{L}_{rec}+\lambda \mathcal{L}_{adv},
\end{equation}
where $\lambda$ is the weight of $\mathcal{L}_{adv}$.
After training, the fingerprint extractor $E$ is frozen when used to extract GAN fingerprints $F$ from fake images by subtracting the reconstructed image from the original fake image: $F={x}_{f}-E({x}_{f})$.

\subsubsection{Fingerprint Perturbation}
GAN-generated image detectors perform poorly on unseen GAN architectures since the GAN fingerprints are architecture-specific.
To address this, we perform random perturbations on the extracted fingerprints to simulate the fingerprints of unknown GANs, so as to boost the generalizability of the GAN detector.
We propose two different perturbation strategies in this paper:

\textbf{Scaling}.
Scaling means the numerical reduction or amplification of the original fingerprint $F$, i.e., multiplying the fingerprint by a random factor $\alpha$. This can be formulated as:
\begin{equation}
\setlength{\abovedisplayskip}{5pt}
\setlength{\belowdisplayskip}{5pt}
    F_{new}={\alpha} F,
\end{equation}
where $\alpha \in \left[-{\alpha}_{0},{\alpha}_{0}\right]$ is the randomly selected scaling factor.

\begin{table*}[t]
    \renewcommand\arraystretch{0.9}
    \centering
    \footnotesize
    \caption{Cross-GAN evaluations.}
    \label{tab:cross-model1}
    % l c r: 左中右对齐
    % \vspace{-0.2cm}
    \begin{tabular}{cccccccccccccc|cc}  
        \toprule
    \multicolumn{1}{c}{\multirow{3}{*}{Method}} & \multicolumn{1}{c}{\multirow{3}{*}{Training Set}} & \multicolumn{12}{c}{Test Sets} \\ \cmidrule{3-16} 
    & & \multicolumn{2}{c}{StyleGAN} & \multicolumn{2}{c}{StyleGAN2} & \multicolumn{2}{c}{BigGAN} & \multicolumn{2}{c}{CycleGAN} & \multicolumn{2}{c}{StarGAN} & \multicolumn{2}{c}{GauGAN} & \multicolumn{2}{|c}{Mean} \\ \cmidrule{3-16} 
    & & Acc & AP & Acc & AP & Acc & AP & Acc & AP & Acc & AP & Acc & AP & Acc & AP \\ 
    \midrule
    Wang~\cite{wang2020cnn} & ProGAN & 71.4 & 96.3 & 67.5 & 93.4 & 60.9 & 83.3 & 83.8 & 94.3 & 84.6 & 93.6 & 79.3 & \textbf{98.1} & 74.6 & 93.2 \\
    Frank~\cite{frank2020leveraging} & ProGAN & 81.8 & 91.7 & 71.4 & 93.0 & 76.0 & 87.8 & 62.8 & 77.3 & 96.9 & 99.4 & 73.9 & 93.1 & 77.1 & 90.4 \\
    Durall~\cite{durall2020watch} & ProGAN & 64.7 & 59.0 & 69.2 & 62.9 & 59.4 & 55.3 & 66.9 & 60.9 & 98.5 & 97.1 & 57.2 & 53.9 & 69.3 & 64.9 \\
    Jeong~\cite{jeong2022bihpf} & ProGAN & 73.0 & 83.9 & 62.7 & 75.9 & 78.1 & \textbf{94.8} & 60.5 & 85.6 & \textbf{100.0} & 100.0 & 68.7 & 97.4 & 73.8 & 89.6 \\
    Jeong~\cite{jeong2022fingerprintnet} & LSUN-horse & 74.1 & 85.3 & \textbf{89.5} & 96.1 & \textbf{85.0} & \textbf{94.8} & 71.2 & \textbf{96.9} & 99.9 & 100.0 & 75.9 & 90.9 & 82.6 & 94.0 \\ \midrule
     \textbf{Ours} -Scaling & ProGAN & \textbf{85.7} & 98.6 & 83.8 & \textbf{98.2} & 81.2 & 85.3 & 83.3 & 93.9 & 99.1 & \textbf{100.0} & 75.1 & 81.3 & \textbf{84.7} & 92.9 \\
    \textbf{Ours} -Mixup & ProGAN & 82.2 & \textbf{98.7} & 78.0 & 98.1 & 79.1 & 84.8 & \textbf{86.4} & 95.6 & 98.8 & \textbf{100.0} & \textbf{83.4} & 90.3 & \textbf{84.7} & \textbf{94.6} \\ \bottomrule
    % \vspace{-0.3cm}
    \end{tabular}
\end{table*}

\textbf{Mixup}.
Mixup means randomly selecting fingerprints from different samples and mixing them with a certain ratio to obtain a new fingerprint.
This can be formulated as:
\begin{equation}
\setlength{\abovedisplayskip}{5pt}
\setlength{\belowdisplayskip}{5pt}
F_{new}=\sum_{i=1}^{n}\left(\beta_{i} F_i\right),
\end{equation}
where $n$ is the number of selected fingerprints, ${\beta}_{i}$ is the mixup ratio of $i$-th fingerprint, and $\sum_{i=1}^{n} {\beta}_{i}=1$.

After the above perturbations, the perturbed fingerprint $F_{new}$ is added back to the reconstructed fake image $E(x_f)$ to obtain the fake image with a new fingerprint.
This process can be formulated as:
\begin{equation}
\setlength{\abovedisplayskip}{5pt}
\setlength{\belowdisplayskip}{5pt}
    {x}_{f}^{new}=E(x_f)+F_{new}.
\end{equation}

\subsection{GAN Detector Training}
In the training of the GAN detector, even if the training data is drawn from a single domain, the types of fingerprints are greatly expanded after the fingerprint perturbation augmentation, resulting in better generalization.
The detector performs a 2-category classification task on real images and perturbed fake images, and it is optimized by binary cross-entropy:
\begin{equation}
\setlength{\abovedisplayskip}{5pt}
\setlength{\belowdisplayskip}{5pt}
    \mathcal{L}_{cls}=-\left[y_{i} \log \left(\hat{y}_{i}\right)+\left(1-y_{i}\right) \log \left(1-\hat{y}_{i}\right)\right],
\end{equation}
where $y_{i} \in\{0,1\}$ denotes the label of the input image,  $\hat{y}_{i} \in [0,1]$ denotes the prediction of the detector.
Again, note that the fingerprint extractor $E$ is frozen during this stage.

% -----------------------------换节分隔符-----------------------------

\section{Experiment}
\subsection{Experimental Settings}
We first introduce the experimental settings, including the dataset, metrics and implementation details in this section.

\textbf{Dataset.}
Our experiments are based on the widely-used \textit{ForenSynths} dataset~\cite{wang2020cnn}.
The training set contains 20 categories of images generated by ProGAN~\cite{karras2018progressive} and the real images in LSUN dataset~\cite{yu2015lsun}.
The test set contains images generated by several well-known GANs (StyleGAN, StyleGAN2, BigGAN, CycleGAN, GauGAN, StarGAN) and the real images used to train them.

\textbf{Evaluation Metrics.}
We use two most used evaluation metrics in GAN forensics, namely Accuracy (Acc) and Average Precision (AP).

\textbf{Implementation Details.}
As in prior works, we employ ResNet50 as the detector to ensure fair comparisons.
We use Adam to optimize the fingerprint extractor, category discriminator, and detector with learning rates of $10^{-3}$, $10^{-3}$, and $10^{-4}$, respectively.
$\lambda$ is set to $10^{-4}$, ${\alpha}_{0}$ and $n$ are set to 5 and 3, respectively.
All images are resized to $256 \times 256$.

\subsection{Cross-Category Evaluations}
Following Jeong \textit{et al.}~\cite{jeong2021self}, we first evaluate our method in a less challenging setting where the detector is trained on a single category of fake images and evaluated on other categories generated by the same GAN.
Specifically, we only use images of category \textit{horse} generated by ProGAN to train the detector and perform evaluations on the images of the remaining 19 categories in the ProGAN dataset.
The results are shown in Table~\ref{tab:cross-cate}, detectors trained with the proposed fingerprint augmentation strategies achieve nearly perfect performances in terms of both Acc and AP.
Our methods outperform the state-of-the-art method by over 4\% in Acc and 1.9\% in AP, demonstrating the effectiveness of both \textbf{Scaling} and \textbf{Mixup} in improving the generalization ability to unseen categories.

\begin{table}[t]
    \footnotesize
    \renewcommand\arraystretch{0.9}
    \centering
    \caption{Comparisons of cross-category performance. The detector is trained on horse images and evaluated on the remaining 19 categories generated by ProGAN.}
    \label{tab:cross-cate}
    % \vspace{-0.2cm}
    \begin{tabular}{cccc}
    \toprule
    Method & Training set & Acc & AP \\ \midrule
    Wang\cite{wang2020cnn} & horse & 50.4 & 63.8 \\
    Frank\cite{frank2020leveraging} & horse & 78.9 & 77.9 \\
    Durall\cite{durall2020watch} & horse & 85.1 & 79.5 \\
    Jeong\cite{jeong2021self} & horse-real & 92.0 & 97.7 \\ \midrule
    \textbf{Ours} -Scaling & horse & \textbf{96.6} & \textbf{99.6} \\
    \textbf{Ours} -Mixup & horse & \textbf{96.0} & \textbf{99.6} \\
    \bottomrule  
    \end{tabular}
    % \vspace{-0.3cm}

\end{table}

\subsection{Cross-GAN Evaluations}
To further demonstrate the generalization of our method to unseen GAN architectures, we perform cross-GAN evaluations where the detector is trained on ProGAN but evaluated on other unseen GANs. 
As shown in Table~\ref{tab:cross-model1}, we present specific results for each unseen GAN and the average performances on all 6 GANs.
For the average performances, the proposed \textbf{Scaling} and \textbf{Mixup} both exceed the state-of-the-art by 2.1\% in Acc.
Although \textbf{Scaling} is slightly below the optimal value in AP, \textbf{Mixup} still maintains the highest level among all detection methods.
The results show that the generalization ability of our augmentation strategies is comparable and outperforms other detection methods by a considerable margin, especially in terms of accuracy.
Compared to Wang \textit{et al.}'s method~\cite{wang2020cnn} that performs data augmentation in the spatial domain such as Gaussian blur and JPEG, our fingerprint domain augmentation has a clear superiority.
Our methods achieve better results on 5 out of 6 GANs and the mean Acc on 6 unseen GANs is 11.1\% higher.
The results show that our method can efficiently improve the generalization of the detector to unseen GANs by perturbing the fingerprints of training images.

\begin{table}[t]
    \renewcommand\arraystretch{0.9}
    \centering
    % \scriptsize
    \footnotesize
    \begin{center}
    \caption{Cross-GAN evaluations with different numbers of categories used in training.}
    % \resizebox{0.7\linewidth}{!}{%
        \centering
        \vspace{-0.2cm}
        \begin{tabular}{cccc}

        \toprule
        Method & Categories & mean Acc & mean AP \\ \midrule
        Wang\cite{wang2020cnn} & 1 & 57.6 & 81.3 \\
        Frank\cite{frank2020leveraging} & 1 & 68.9 & 74.9 \\
        Durall\cite{durall2020watch} & 1 & 68.6 & 63.8 \\
        Jeong\cite{jeong2022bihpf} & 1 & 75.2 & 84.7 \\
        \textbf{Ours} -Scaling & 1 & \textbf{79.8} & \textbf{86.0} \\
        \textbf{Ours} -Mixup & 1 & 76.8 & \textbf{86.0} \\ 
        \midrule
        Wang\cite{wang2020cnn} & 2 & 57.3 & 82.1 \\
        Frank\cite{frank2020leveraging} & 2 & 77.1 & 84.7 \\
        Durall\cite{durall2020watch} & 2 & 66.4 & 61.5 \\
        Jeong\cite{jeong2022bihpf} & 2 & 80.1 & 82.3 \\
        \textbf{Ours} -Scaling & 2 & 78.9 & 90.3 \\
        \textbf{Ours} -Mixup  & 2 & \textbf{81.4} & \textbf{91.0} \\ 
        \midrule
        Wang\cite{wang2020cnn} & 4 & 65.6 & 89.0 \\
        Frank\cite{frank2020leveraging} & 4 & 68.9 & 83.4 \\
        Durall\cite{durall2020watch} & 4 & 69.5 & 65.1 \\
        Jeong\cite{jeong2022bihpf} & 4 & 80.0 & 83.1 \\
        \textbf{Ours} -Scaling & 4 & \textbf{81.7} & \textbf{91.0} \\
        \textbf{Ours} -Mixup & 4 & 81.5 & 90.5 \\ 
        \bottomrule
        \end{tabular}
    % }
    
    \label{tab:cross-model2}
    \end{center}
    \vspace{-0.3cm}
\end{table}

In addition, we further evaluate the effect of the number of categories in the training set on the cross-GAN performances.
Following Jeong \textit{et al.}~\cite{jeong2022bihpf}, the categories used in the training of the detector are \textit{horse}, \textit{car}, \textit{cat}, and \textit{airplane}, with the number of categories varying from 1 to 4.
As shown in Table~\ref{tab:cross-model2}, the mean Acc and mean AP on the other 6 unseen GANs are presented.
It is obvious that our methods achieve the highest generalizability when trained on only a few categories compared to other methods.
Our method only takes a small number of categories to obtain good generalization, and as the number of categories increases, the generalization can be increased further.

\subsection{Visualizations}
\begin{figure}[t]
\centering
\includegraphics[width=0.9\linewidth]{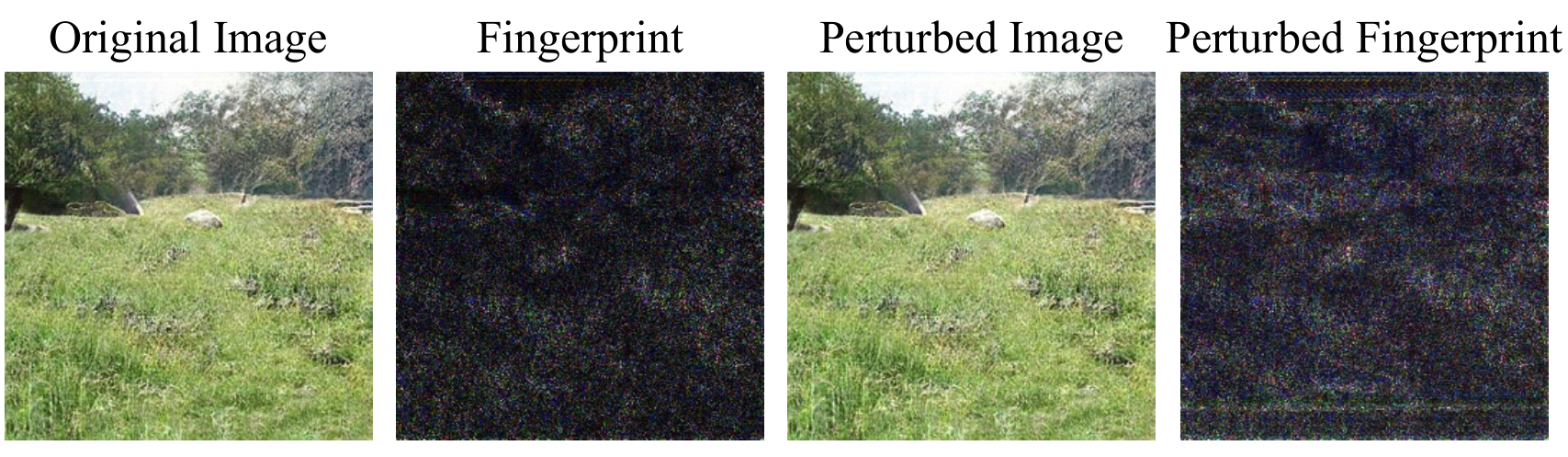}
\caption{Visualizations of fingerprint perturbation.}
\label{fig:vis}
\end{figure}

In Fig.~\ref{fig:vis}, we present an image generated by ProGAN and its perturbed version using \textbf{Mixup}.
The two images are visually almost indistinguishable while their fingerprints have a significant difference.
The perturbation successfully simulates unknown types of GANs, allowing the detector to be trained on more kinds of fingerprints.

\begin{figure*}[t]
\centering
\includegraphics[width=0.92\linewidth]{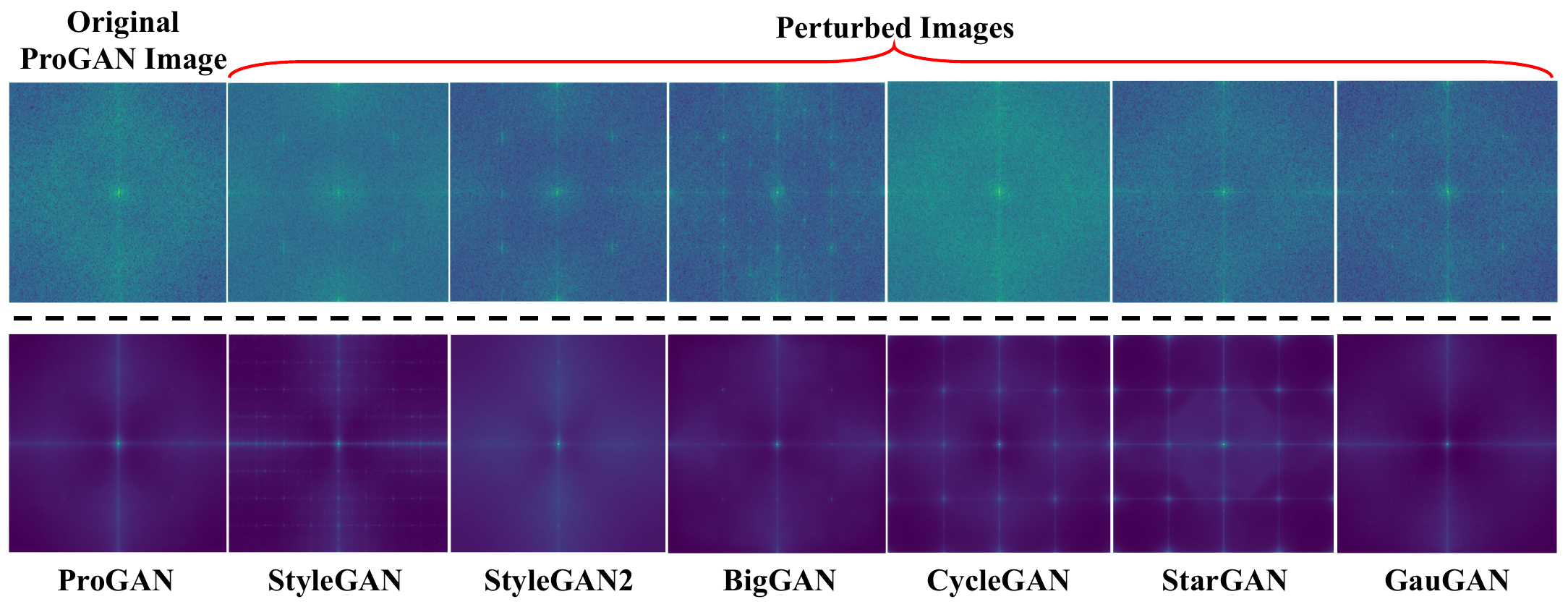}
\caption{Visualizations of the spectra for perturbed images and different GANs.}
\label{fig:spectrum}
\end{figure*}

In Fig.~\ref{fig:spectrum}, we show the spectra of images generated from different GANs and the spectra of perturbed ProGAN images.
The first row shows the spectra of a high-pass filtered original ProGAN image and the corresponding perturbed images using the proposed fingerprint augmentation strategies.
The second row shows the averaged spectra of 2000 high-pass filtered images generated by different GANs.
We can see that the fingerprint perturbation has an obvious effect on the spectra.
The spectra of the perturbed images exhibit distinct patterns from the original image.
Although fingerprint perturbation does not reproduce the spectra of unseen GANs, the training domain is significantly extended.
This also corroborates the advantage of our method in generalizability.

\subsection{Ablation Studies}
To demonstrate the effectiveness of the category discriminator, we perform ablation studies on $\mathcal{L}_{adv}$ under the cross-GAN evaluations in Table~\ref{tab:cross-model1}.
The results are presented in Table~\ref{tab:ablation-adv}, which shows that with $\mathcal{L}_{adv}$, both \textbf{Scaling} and \textbf{Mixup} are 
slightly increased in mean Acc and AP.
This is probably due to the fact that the fingerprint extractor trained with $\mathcal{L}_{adv}$ can extract fingerprints more precisely, which is advantageous for the detector that relies on the GAN fingerprints.

\begin{table}[t]
    \renewcommand\arraystretch{0.9}
    \centering
    \footnotesize

    \begin{center}
    \caption{Effectiveness of the category discriminator.}
    % \resizebox{0.6\linewidth}{!}{%
        \centering
        \vspace{-0.2cm}
        \begin{tabular}{cccc}
        
        \toprule
        Method & $\mathcal{L}_{adv}$ & mean Acc & mean AP \\ \midrule
        \textbf{Scaling}  & - & 83.1 & 92.5 \\
        \textbf{Scaling}  & $\checkmark$ & \textbf{84.7} & \textbf{92.9} \\ \midrule
        \textbf{Mixup}  & - & 84.3 & 93.0 \\
        \textbf{Mixup}  & $\checkmark$ & \textbf{84.7} & \textbf{94.6} \\\bottomrule

        \end{tabular}
    % }
    \vspace{-0.3cm}
    \label{tab:ablation-adv}
    \end{center}
\end{table}

\begin{table}[t!]
    \renewcommand\arraystretch{0.9}
    \centering
    \footnotesize

    \begin{center}
    \caption{Effectiveness of different detectors.}
    % \resizebox{0.8\linewidth}{!}{%
        \centering
        \vspace{-0.2cm}
        \begin{tabular}{ccc}
        
        \toprule
        Method & mean Acc & mean AP \\ \midrule
        \textbf{Scaling} -ResNet18 & 84.3 & 91.4 \\
        \textbf{Mixup} -ResNet18 & \textbf{85.5} & \textbf{91.9} \\ \midrule
        \textbf{Scaling} -Xception & \textbf{87.2} & 95.8 \\
        \textbf{Mixup} -Xception & 86.9 & \textbf{96.1} \\ \bottomrule
        
        \end{tabular}
    % }
    \vspace{-0.3cm}
    \label{tab:ablation-classifiers}
    \end{center}
\end{table}

\begin{table}[t!]
    \renewcommand\arraystretch{0.9}
    \centering
    \footnotesize

    \begin{center}
    \caption{Ablation studies with various parameters.}
    % \resizebox{0.6\linewidth}{!}{%
        \centering
        \vspace{-0.2cm}
        \begin{tabular}{cccc}
        
        \toprule
        Method & Params & mean Acc & mean AP \\ \midrule
        \textbf{Scaling} & ${\alpha}_{0}$=1 & 84.7 & 92.2 \\
        \textbf{Scaling} & ${\alpha}_{0}$=2 & \textbf{85.3} & 91.6 \\
        \textbf{Scaling} & ${\alpha}_{0}$=5 & 84.7 & \textbf{92.9} \\
        \textbf{Scaling} & ${\alpha}_{0}$=10 & 83.7 & 92.1 \\
        \textbf{Scaling} & ${\alpha}_{0}$=100 & 82.6 & 92.2 \\ \midrule
        \textbf{Mixup} & $n$=2 & \textbf{86.1} & 91.7 \\
        \textbf{Mixup} & $n$=3 & 84.7 & \textbf{94.6} \\
        \textbf{Mixup} & $n$=5 & 84.1 & 91.9 \\
        \textbf{Mixup} & $n$=10 & 83.3 & 91.0 \\  \bottomrule

        \end{tabular}
    % }
    \vspace{-0.3cm}
    \label{tab:various-params}
    \end{center}
\end{table}

Then we evaluate the effectiveness of different detectors.
Given that we used ResNet50 in previous experiments, we additionally use 2 different architectures ResNet18 and Xception as the GAN detector.
As shown in Table~\ref{tab:ablation-classifiers}, the mean AP of Xception detector is 3\% to 4\% higher than that of ResNet18 and ResNet50, and the mean Acc only shows moderate differences.
This suggests that the proposed method is effective for detectors of different architectures.

Table~\ref{tab:various-params} presents the results of different $\alpha_{0}$ and $n$ for \textbf{Scaling} and \textbf{Mixup}.
It is apparent that the optimal accuracy is attained when these two parameters are set to moderate values, as excessively large values have a notable adverse effect on the accuracy.

% -----------------------------换节分隔符-----------------------------

\section{Conclusion}
In this work, we propose a novel fingerprint domain augmentation method for general GAN-generated image detection.
Unlike previous methods that perform data augmentation in the spatial domain, we directly augments fake images in the fingerprint domain to enrich the fake types of training data.
Extensive experiments demonstrate the effectiveness of the fingerprint domain augmentation and its excellent generalizability compared to the state-of-the-art works.

\section*{Acknowledgment}
This work is supported in part by the National Key R\&D Program of China under Grant numbers 2022YFB3103100, 2020YFB1005600, in part by Guangdong Basic and Applied Basic Research Foundation under Grant number 2019B1515120010, in part by the National Natural Science Foundation of China under Grant numbers 62122032, 62172233, 62102189, U1936118, 61932011, 61931004, 61825203, U1736203, and 61732021, in part by the Major Program of Guangdong Basic and Applied Research Project under Grant number 2019B030302008.

\bibliographystyle{IEEEtran}
\bibliography{ICME2023}

\end{document}